\pdfoutput=1

\documentclass[11pt]{article}

\usepackage{acl}

\usepackage{times}
\usepackage{latexsym}

\usepackage[T1]{fontenc}

\usepackage[utf8]{inputenc}

\usepackage{microtype}

\usepackage{inconsolata}

\usepackage{bm}
\usepackage{tikz}
\usepackage{array}
\usepackage{float}
\usepackage{bbding}
\usepackage{xspace}
\usepackage{amssymb}
\usepackage{amsmath}
\usepackage{enumitem}
\usepackage{graphicx}
\usepackage{multirow}
\usepackage{multicol}
\usepackage{makecell}
\usepackage{listings}
\usepackage{setspace}
\usepackage{booktabs}
\usepackage{hyperref}
\usepackage{subfigure}
\usepackage[most]{tcolorbox}
\usepackage[ruled]{algorithm2e}

\setlist[itemize]{leftmargin=5mm, itemsep=0mm}

\newcommand{\ie}{\emph{i.e.,}\xspace}
\newcommand{\eg}{\emph{e.g.,}\xspace}

\newcommand{\aka}{\emph{a.k.a.,}\xspace}

\newcommand{\name}{StructureCoder\xspace}

\title{Alignment with Fill-In-the-Middle for Enhancing Code Generation}

\author{
  Houxing Ren$^1$ \quad Zimu Lu$^1$ \quad Weikang Shi$^1$ \quad Haotian Hou$^{2,5}$ \quad Yunqiao Yang$^1$  \\
  \textbf{Ke Wang}$^1$ \quad \textbf{Aojun Zhou}$^1$ \quad \textbf{Junting Pan}$^{1,3}$ \quad \textbf{Mingjie Zhan}$^2\footnotemark[1]$ \quad \textbf{Hongsheng Li}$^{1,3,4}$\thanks{Corresponding author.} \\
  $^1$CUHK MMLab ~ $^2$SenseTime Research ~ $^3$CPII under InnoHK \\ $^4$Shanghai AI Laboratory ~ $^5$Beihang University \\
  renhouxing@gmail.com \quad zhanmingjie@sensetime.com \quad hsli@ee.cuhk.edu.hk
}

\begin{document}

\maketitle

\begin{abstract}
The code generation capabilities of Large Language Models (LLMs) have advanced applications like tool invocation and problem-solving. However, improving performance in code-related tasks remains challenging due to limited training data that can be verified with accurate test cases. While Direct Preference Optimization (DPO) has shown promise, existing methods for generating test cases still face limitations. In this paper, we propose a novel approach that splits code snippets into smaller, granular blocks, creating more diverse DPO pairs from the same test cases. Additionally, we introduce the Abstract Syntax Tree (AST) splitting and curriculum training method to enhance the DPO training. Our approach demonstrates significant improvements in code generation tasks, as validated by experiments on benchmark datasets such as HumanEval (+), MBPP (+), APPS, LiveCodeBench, and BigCodeBench. Code and data are available at \url{https://github.com/SenseLLM/StructureCoder}.
\end{abstract}

\section{Introduction} \label{sec:intro}

Large Language Models (LLMs) have significantly enhanced applications like tool invocation and mathematical problem-solving~\cite{GPT42023ABS230308774, LLama22023ABS230709288, Mixtral2023ABS240104088, Qwen252024Yang}. To improve LLM performance, a common approach involves two stages: supervised fine-tuning~(SFT) and alignment. The alignment phase is particularly effective, with Direct Preference Optimization~(DPO)~\cite{DPO2023Rafailov} gaining popularity for its simplicity and effectiveness. DPO has been successfully applied in fields like dialogue~\cite{Llama32024Dubey}, faithfulness~\cite{ContextDPO2024Bi}, and reasoning~\cite{SDPO2024Lai}.

However, despite its success in several domains, DPO has shown limited improvement in code generation tasks, and in some cases, its application may even be counterproductive~\cite{DPOAnalysis2024Xu}.
This limitation may stem from the scarcity of training data containing test cases~\cite{PLUM2024Zhang}, as seen in training sets like APPS~\cite{APPS2021Dan}, which include only 5,000 samples.
To address this, PLUM~\cite{PLUM2024Zhang} and CodeDPO~\cite{CodeDPO2024Zhang} have proposed methods for constructing test cases to expand the training dataset for DPO. 
However, these approaches still face criticism for their inability to fully guarantee the accuracy of test samples, which limits the achievable performance improvements.

To overcome these challenges, we propose a novel approach that makes more efficient use of the limited, high-quality data available. Inspired by recent advancements in the mathematics domain~\cite{LVSBS2024Lightman, SCDPO2024Lu, lu2024mathgeniegeneratingsyntheticdata, SDPO2024Lai}, we decompose code snippets into smaller, more granular blocks. This allows for the construction of more detailed and diverse DPO pairs. By treating each block as a target for prediction, the model can fine-tune on smaller code segments, enabling more effective learning and potentially improving performance in code generation tasks. Furthermore, this approach generates a larger number of DPO pairs, making better use of the available training set.

\begin{figure*}[t]
    \centering
    \includegraphics[width=\textwidth]{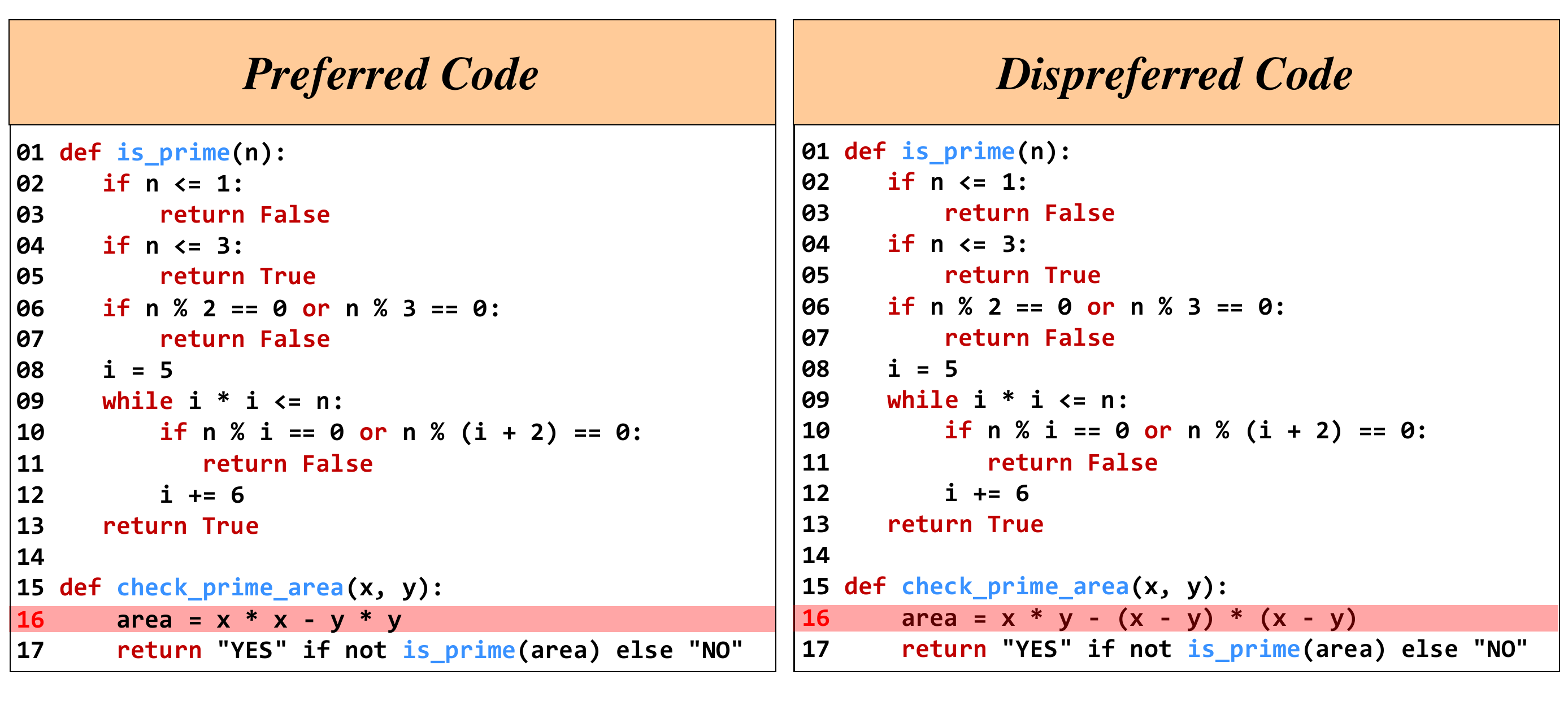}
    \caption{A preference pair case in the code generation field. The left is the correct response, and the right is the incorrect response. The only difference between the two responses is in Line 16.}
    \label{fig:intro}
\end{figure*}

Building on these insights, we introduce a novel method called \name, which focuses on maximizing the utility of limited training data. Specifically, we leverage the fill-in-the-middle~(FIM) approach~\cite{FIM2022ABS220714255}, an important capability of code LLMs for code completion, to generate fine-grained DPO pairs. By using FIM, we can divide a code snippet into multiple blocks by Abstract Syntax Tree~(AST), each serving as a target, and then prompt the model to generate the missing block. Test cases are used to evaluate the correctness of the generated block, thereby constructing accurate DPO pairs. Additionally, we propose a curriculum training method that organizes the training set according to the depth of the target block. This method progressively increases training difficulty, leading to better performance.

Our contributions can be summarized as follows: 
\begin{itemize} 
    \item We propose the use of FIM to enhance DPO for code LLMs, enabling the effective use of limited training data with test cases. 
    \item We introduce a method for generating accurate FIM data through AST-based block segmentation, and a curriculum training strategy to organize the training set by target block depth, leading to improved performance.
    \item We conduct extensive experiments on HumanEval~(+), MBPP~(+), APPS, LiveCodeBench, and BigCodeBench to demonstrate the effectiveness of the proposed method in code-related tasks. 
\end{itemize}
\section{Preliminaries} \label{sec:pre}

In this section, we introduce the fundamental components of DPO and FIM, followed by an analysis of the limitations of DPO in code generation tasks.

\subsection{Direct Preference Optimization}

DPO~\cite{DPO2023Rafailov} offers a solution that bypasses reward model training, instead directly fine-tuning the LLM using preference pairs. The DPO loss is defined as
\begin{equation*} \small
    \begin{aligned}
        &\mathcal{L}_{DPO}(y_w, y_l, x) = \\
        &- \log \sigma\left(\beta \log \frac{\pi_\theta(y_w \mid x)}{\pi_{ref}(y_w \mid x)} - \beta \log \frac{\pi_\theta( y_l \mid x)}{\pi_{ref}(y_l \mid x)}\right),
    \end{aligned}
\end{equation*}
where $y_w$ is the preferred response and $y_l$ is the dispreferred response.

\subsection{Fill-In-the-Middle~(FIM)} \label{appendix:fim}

When performing FIM~\cite{FIM2022ABS220714255} training, we swap the middle and the suffix segments. Specifically, after splitting, it moves the suffix segment before the middle segment:
\begin{equation*}
    \text{code} \to \text{(pre, mid, suf)} \to \text{(pre, suf, mid)}.
\end{equation*}
Then concatenate the three pieces using special tokens as
\begin{equation*}
    \text{<PRE> pre <SUF> suf <MID> mid <EOT>}.
\end{equation*}
After that, we use the format to train a causal language model and use the prefix and suffix segments to predict the middle segment in the inference stage.

\subsection{Weakness of DPO} \label{sec:weak_of_dpo}

Prior work~\cite{QStar2024Rafailov} demonstrates that DPO can implicitly learn token-level reward functions within the Markov Decision Process framework of large language models. This enables DPO-trained models to assign differentiated rewards to individual tokens, effectively identifying those associated with factual inaccuracies while maintaining consistent rewards elsewhere. However, accurately capturing such fine-grained reward signals requires substantial training data, as the quality and scale of the dataset critically influence the learned reward and the performance.

This ability introduces two challenges for DPO in the code field. On the one hand, there are limited training sets containing test cases. On the other hand, the preferred and dispreferred responses often share a similar structure, with only minimal differences in detailed expression. This phenomenon causes DPO to need a larger training data set to learn token-level rewards. This is because when constructing DPO training data, a model is used to generate multiple responses, which are then evaluated against test cases. This method often results in correct responses and incorrect responses being nearly identical, differing only in small details such as a specific expression, an if-block, or a function.

To illustrate the consequences of this scenario, we present an extreme example, as shown in Figure~\ref{fig:intro}. In this case, we assume that only a single segment differs between a correct response $y_w$ and an incorrect response $y_l$. These sequences are defined as
\begin{align*}
    y_w &= \{\text{pre}, \text{mid}_w, \text{suf}\}, \\
    y_l &= \{\text{pre}, \text{mid}_l, \text{suf}\}.
\end{align*}
Here, the prefix is the same for both sequences, so the loss for the prefix is zero. The DPO loss can then be expressed as
\begin{align*}
    \mathcal{L}(y_w, y_l, x) = \mathcal{L}_{DPO}(\text{mid}_w | \text{suf}, \text{mid}_l | \text{suf}, x | \text{pre}),
\end{align*}
where $|$ denotes concatenation. Let $x_p = x | \text{pre}$, $x_w = x | \text{pre} | \text{mid}_w$, and $x_l = x | \text{pre} | \text{mid}_l$. The argument of the log sigmoid function in the loss can then be separated into two components:
\begin{equation*}
    \beta \log \frac{\pi_\theta(\text{mid}_w \mid x_p)}{\pi_{ref}(\text{mid}_w \mid x_p)} - \beta \log \frac{\pi_\theta(\text{mid}_l \mid x_p)}{\pi_{ref}(\text{mid}_l \mid x_p)},
\end{equation*}
\begin{equation*}
    \beta \log \frac{\pi_\theta(\text{suf} \mid x_w)}{\pi_{ref}(\text{suf} \mid x_w)} - \beta \log \frac{\pi_\theta(\text{suf} \mid x_l)}{\pi_{ref}(\text{suf} \mid x_l)}.
\end{equation*}
The first term corresponds to the loss for the middle segment, which is the key component of the overall loss. The second term, however, has a negative impact, as it implies the model should not generate the suffix based on $x_l$. However, this is an issue because a prior error is unrelated to the rest segment. For instance, as shown in Figure~\ref{fig:intro}, even though there's an error in Line 16, the remaining function should still be generated like this. 

As demonstrated, it is inappropriate to calculate the DPO loss on the suffix, as it leads to negative rewards for correctly generated tokens. Unfortunately, only a small fraction of the generated code typically contains errors; the majority is correct. This implies the need for a large dataset with test cases to make up for the negative impact of the DPO loss on the suffix. 

\begin{figure*}[t]
    \centering
    \includegraphics[width=\textwidth]{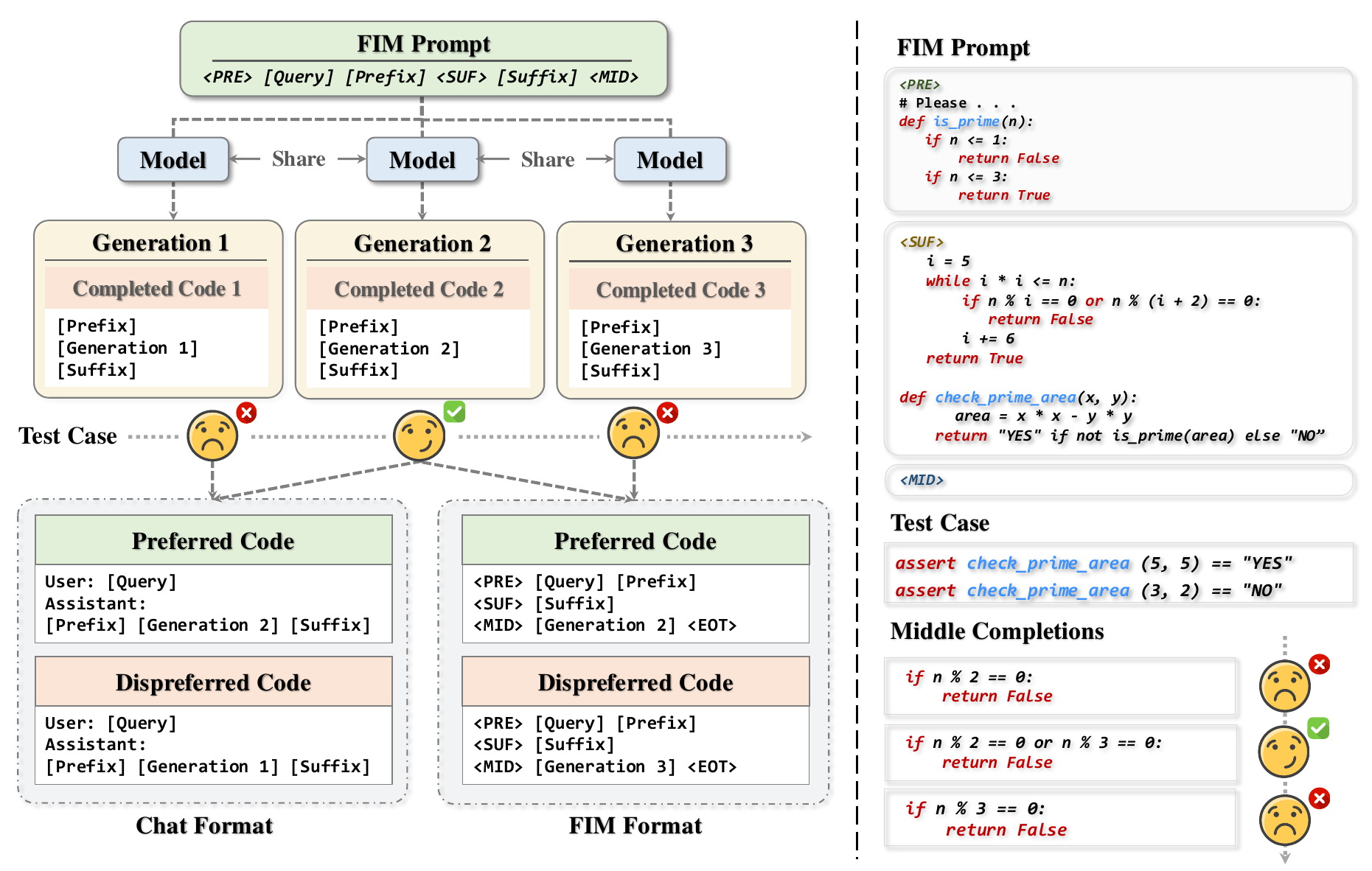}
    \caption{An overview of our FIM-style preference modeling process. A concrete example (right) illustrates the task of completing an is\_prime function, with correctness judged via downstream function tests.}
    \label{fig:fim_dpo}
\end{figure*}

\section{Methodology} \label{sec:met}

In Section~\ref{sec:pre}, we analyze the impediments of DPO with limited code training data. In this section, we introduce a novel approach that effectively mitigates these weaknesses. Unlike previous methods~\cite{PLUM2024Zhang, CodeDPO2024Zhang}, which focus on expanding the training dataset, we emphasize optimizing the utilization of the limited available data. We begin by demonstrating how to leverage FIM to avoid the detrimental part of DPO loss. Next, we describe how to construct more accurate FIM data. Finally, we provide an overview of the entire training pipeline.

\subsection{FIM Enhanced DPO}

To alleviate the negative impact of the DPO loss on the suffix part, we proposed to combine FIM and DPO to enhance the code generation performance.
The FIM ability of code LLMs allows them to generate the middle segment based on the prefix and suffix segments, enabling us to control what is included in the model response flexibly.

As shown in Figure~\ref{fig:fim_dpo}, given a code training case, which includes a code problem $q$, a reference solution $c$ to solve the problem, and several test cases $t$ to test any generation, we first select consequent snippets from the golden code $n$ times to construct the target snippets $M = \{m_1, m_2, \dots, m_n\}$. This process also generate the corresponding prefixes $P = \{p_1, p_2, \dots, p_n\}$ and suffixes $S = \{s_1, s_2, \dots, s_n\}$. Then we construct the FIM prompt with the target snippets, prefixes, and suffixes:
\begin{equation} \label{eq:fim_format}
    \text{<PRE>} ~ \text{Convert}(q) ~ p_i ~ \text{<SUF>} ~ s_i ~ \text{<MID>},
\end{equation}
where ``Convert'' denotes a function converting the text prompts into comments in the code. Then, we task the code LLM to generate $m$ generations for each prefix and suffix pair: $G = \{g_1^{(1)}, \dots, g_1^{(m)}, \dots, g_n^{(1)}, \dots, g_n^{(m)},\}$ based on the prompt. Finally, we concatenate the generation with the corresponding prefix and suffix, \eg $p_1 \mid g_1^{(1)} \mid s_1$, and use the test cases $t$ to verify the generation.

For each target snippet, we select a correct response for each incorrect response, based on the minimum edit distance, to construct the preference pair. Then we apply two prompt formats to fine-tune the model. One is the FIM format:
\begin{equation}
    \text{<PRE>} ~ \text{Convert}(q) ~ p_i ~ \text{<SUF>} ~ s_i ~ \text{<MID>} ~ g_i^{(j)}.
\end{equation}
Here, only $g_i^{(j)}$ is regarded as the response and included in the DPO loss.
Another format is the Chat format:
\begin{equation} \label{eq:chat_format}
    \text{User:} ~ q. ~~ \text{Assistant:} ~ p_i ~ g_i^{(j)} ~ s_i.
\end{equation}
Also, only $g_i^{(j)}$ is calculated in the DPO loss. This format is designed to preserve the model's chat capabilities. During training, we randomly choose one of two formats for each sample by drawing from a Bernoulli distribution with probability $\alpha$, which determines the proportion of samples using each format. This strategy enables the model to be fine-tuned on shorter code snippets, thereby improving learning efficiency.

\begin{figure}[t]
    \centering
    \includegraphics[width=\columnwidth]{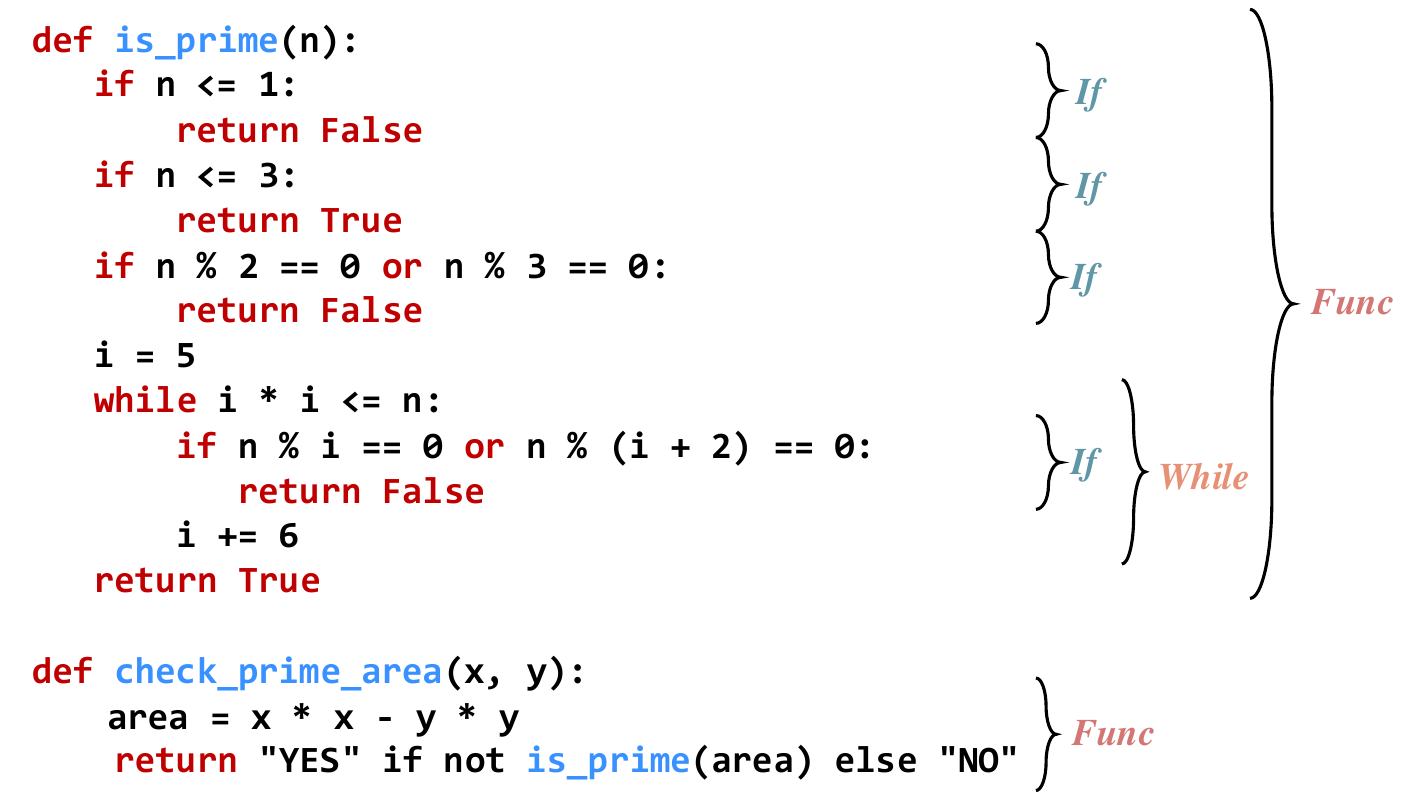}
    \caption{Illustration of our AST-based segmentation strategy. This segmentation ensures each middle segment is syntactically and semantically coherent, enabling more effective FIM-based fine-tuning.}
    \label{fig:ast}
\end{figure}

\subsection{AST Enhanced FIM}

As discussed in Section~\ref{sec:pre}, errors in the middle segment are independent of those in the suffix segment. Therefore, it is essential to select appropriate consequent code snippets from the golden label. The ideal snippets should have the following characteristics to ensure effective model learning and improved performance:
\begin{itemize}
    \item The selected code snippets should be distinct blocks to maintain clear boundaries between different functional units, ensuring that each segment functions independently without relying on adjacent ones. 
    \item The selected code snippets should encompass a variety of code structures to enable the model to learn a broad range of coding patterns.
\end{itemize}

Inspired by previous work~\cite{ASTT52024ABS240103003}, we parse a code into an Abstract Syntax Tree~(AST). As illustrated in Figure~\ref{fig:ast}, we first parse the golden label into an AST, then select several blocks as the target code snippets. In this approach, we consider only four node types in the AST: \ie if block, for block, while block, and function block.

By analyzing the AST, we can extract syntactically complete and functionally independent code blocks, which reduces the risk of including partial or entangled code, ensures that the middle segments are self-contained and coherent, and exposes the model to a broader range of distinct structural patterns. For instance, as shown in Figure~\ref{fig:ast}, the first three if blocks handle distinct conditions and thus operate independently, focusing solely on their respective logic and ignoring others. Additionally, the four selected node types represent the core building blocks of Python code, capturing a diverse range of structures commonly encountered in Python programming.

\subsection{Training and Discussion}

To effectively leverage all the blocks selected from AST, we propose a curriculum learning strategy~\cite{CurriculumLearning2009BengioLCW09} that orders the training data based on the length of code snippets. The goal is to enable the model to first focus on mastering token-level rewards for shorter, simpler code snippets before gradually progressing to longer, more complex ones, and finally the whole codes. This strategy leverages the inherent structure of code, where shorter snippets tend to contain simpler constructs and fewer dependencies, making them easier to learn at the token level.

Specifically, we first sort the training samples by the number of lines in the corresponding code snippets. During the initial training stages, the model is fine-tuned on shorter code snippets, which typically have fewer tokens and simpler logic. This allows the model to learn the fundamental code patterns without being overwhelmed by the complexity of longer code blocks. As the training procedure, we progressively introduce longer snippets, which feature more intricate logic, dependencies, and structures.

In summary, the training begins by selecting target code snippets by AST segmentation from the golden code, using FIM to generate middle segments based on the prefix and suffix. Then the training data is ordered by code snippet length, with the model first learning from shorter, simpler code and progressively moving to longer, more complex snippets, ensuring a structured and efficient learning process. Finally, we fine-tune the code LLMs with the sorted training data and randomly select the prompt format (FIM format or chat format) by drawing from a Bernoulli distribution with probability $\alpha$ in each training step.

\section{Experiments} \label{sec:exp}

\begin{table*}[t]
\centering
\setlength{\tabcolsep}{4pt}
\begin{tabular}{l|cc|cc|c|c|cc}
\toprule
\multirow{2.5}{*}{Method} & \multicolumn{2}{c|}{HumanEval} & \multicolumn{2}{c|}{MBPP} & \multirow{2.5}{*}{APPS} & \multirow{2.5}{*}{LiveCodeBench} & \multicolumn{2}{c}{BigCodeBench} \\ \cmidrule(lr){2-5} \cmidrule(lr){8-9}
~ & Base & Plus & Base & Plus & ~ & ~ & Full & Hard \\
\midrule \multicolumn{9}{c}{Closed-Source Models} \\ \midrule
Claude-3.5-Sonnet & 92.1 & 86.0 & 91.0 & 74.6 & - & 37.1 & 45.3 & 23.6 \\
GPT-4o & 92.1 & 86.0 & 86.8 & 72.5 & - & 33.0 & 50.1 & 25.0 \\
o1-mini & 97.6 & 90.2 & 93.9 & 78.3 & - & 54.1 & 46.3 & 23.0 \\
o1-preview & 95.1 & 88.4 & 93.4 & 77.8 & - & 42.5 & 49.3 & 27.7 \\
\midrule \multicolumn{9}{c}{Open-Source Models} \\ \midrule
DS-Coder-1.3B & 65.9 & 60.4 & 65.3 & 54.8 & \underline{5.7} & \underline{4.1} & 22.8 & 3.4 \\
Qwen2.5-Coder-1.5B & 70.7 & 66.5 & 69.2 & 59.4 & \underline{7.7} & \underline{6.7} & \underline{25.2} & \underline{4.1} \\
~~ w/ DPO & 74.4 & 69.5 & 73.5 & 63.8 & 9.3 & 7.8 & 28.6 & 6.8 \\
~~ w/ KTO & 74.4 & 70.1 & 74.1 & 62.4 & 9.4 & 9.4 & 28.0 & 7.4 \\
~~ w/ Focused-DPO & 72.6 & 68.3 & 72.8 & 62.7 & 9.2 & 9.1 & \textbf{29.2} & 7.4 \\
~~ w/ \name & \textbf{75.6}$^{\ddagger}$ & \textbf{71.3}$^{\ddagger}$ & \textbf{75.7}$^{\ast\dagger\ddagger}$ & \textbf{64.8}$^{\ast\dagger\ddagger}$ & \textbf{9.9}$^{\ast\dagger\ddagger}$ & \textbf{10.8}$^{\ast\ddagger}$ & \textbf{29.2}$^{\ast\dagger}$ & \textbf{8.1}$^{\ast\ddagger}$ \\ \midrule
Qwen2.5-Coder-3B & 84.1 & 80.5 & 73.6 & 62.4 & \underline{10.9} & \underline{10.4} & 35.8 & 14.2 \\
~~ w/ DPO & 84.8 & 80.5 & 75.4 & 64.0 & 12.7 & 10.8 & 37.4 & 12.8 \\
~~ w/ KTO & 81.1 & 77.4 & 73.3 & 63.0 & 12.4 & 10.8 & 36.6 & 11.5 \\
~~ w/ Focused-DPO & \textbf{86.0} & 81.1 & \textbf{75.7} & 64.3 & \textbf{13.9} & 10.8 & 37.9 & 14.5 \\
~~ w/ \name & \textbf{86.0}$^{\ast\dagger}$ & \textbf{82.3}$^{\ast\dagger}$ & \textbf{75.7}$^{\ast\dagger}$ & \textbf{64.6}$^{\ast\dagger\ddagger}$ & \textbf{13.9}$^{\ast\dagger}$ & \textbf{13.4}$^{\ast\dagger\ddagger}$ & \textbf{38.4}$^{\ast\dagger\ddagger}$ & \textbf{16.9}$^{\ast\dagger\ddagger}$ \\ \midrule
CodeLlama-7B & 40.9 & 33.5 & 54.0 & 44.4 & \underline{4.3} & \underline{7.1} & 21.9 & 3.4 \\
DS-Coder-6.7B & 74.4 & 71.3 & 74.9 & 65.6 & \underline{8.5} & \underline{10.4} & 35.5 & 10.1 \\
Qwen2.5-Coder-7B & 88.4 & 84.1 & 83.5 & 71.7 & \underline{16.2} & \underline{16.8} & 41.0 & 18.2 \\
~~ w/ DPO & 88.4 & 84.1 & 83.1 & 71.4 & 19.8 & 17.5 & \textbf{41.8} & 19.1 \\
~~ w/ KTO & 89.0 & 84.1 & 85.4 & 72.5 & 19.0 & 16.9 & 41.0 & 18.2 \\
~~ w/ Focused-DPO & 89.0 & 84.8 & 85.4 & 72.5 & 20.0 & 17.9 & 41.1 & 18.2 \\
~~ w/ \name & \textbf{90.9}$^{\ast\dagger\ddagger}$ & \textbf{87.2}$^{\ast\dagger\ddagger}$ & \textbf{85.7}$^{\ast}$ & \textbf{74.3}$^{\ast\dagger}$ & \textbf{20.1}$^{\ast\dagger}$ & \textbf{18.3}$^{\ast\dagger\ddagger}$ & \textbf{41.8}$^{\dagger}$ & \textbf{19.6}$^{\ast\dagger\ddagger}$ \\ \bottomrule
\end{tabular}
\caption{Pass@1 accuracy on HumanEval~(+), MBPP~(+), APPS, LiveCodeBench~(July 2024 - January 2025), and BigCodeBench. The best results of each base are in bold, and results unavailable are left blank. The results re-evaluated on our end are marked with an underline. $\ast$, $\dagger$, and $\ddagger$ denote that our method significantly outperforms DPO, KTO, and Focused-DPO at the level of 0.05, respectively. Note that all results of the LiveCodeBench have been re-evaluated due to the dataset update. We also report the re-evaluated results of the BigCodeBench of Qwen2.5-Coder-1.5B-Inst because the reproduced results differed significantly from the original paper.}
\label{tab:main}
\end{table*}

In this section, we construct extensive experiments to demonstrate the effectiveness of the proposed method and analyze the proposed method.

\subsection{Experimental Setup} \label{sec:setup}

\paragraph{Training Dataset.} Our training dataset is the APPS training set~\cite{APPS2021Dan}, which is collected from different open-access coding websites such as Codeforces, Kattis, and more. The training set includes 5,000 samples and each sample contains a query, several golden labels, and several test cases.

\paragraph{Test Dataset.}
We evaluate our method on HumanEval~\cite{CodexABS210703374} and MBPP~\cite{MBPP2021ABS210807732}, two of the most widely used
benchmarks for code generation.
Considering the insufficiency of test cases in these benchmarks, 
\citet{EvalPlus2023LiuXW023} proposed HumanEval+ and MBPP+, which contain 80×/35× more tests. We also evaluate our method on APPS~\cite{APPS2021Dan} and LiveCodeBench~\cite{LiveCodeBench2024Naman}, two datasets with more difficult code problems. In addition, we evaluate our method on BigCodeBench~\cite{BigCodeBench2024Terry}, which challenged LLMs to invoke multiple function calls as tools.

\paragraph{Implementation Details.} 
We test our methods on Qwen2.5-Coder-Instruct 1.5B/3B/7B~\cite{QwenCoder2024Binyuan}.
We first use Black\footnote{\url{https://github.com/psf/black}} to format all golden labels in the training set, and then use AST\footnote{\url{https://docs.python.org/3/library/ast.html}} to parse the formatted code and extract all desired blocks, \ie if block, for block, while block, function block. Then we use the code LLMs to generate 5 responses based on the FIM prompt, with top\_p = 0.95 and temperature = 0.7. After that, we use the test cases to evaluate the generated responses and select one preferred response and one dispreferred response (if there is no correct generation or no incorrect generation, we will discard this case). 
For training, we fine-tune all models for 3 epochs. We employ an RMSProp optimizer~\cite{RMSProp2013Graves} with a learning rate of 1e-6, a 0.05 warm-up ratio, and a cosine scheduler. We set the batch size as 128 and the max sequence length as 2048. We set the hyperparameter $\alpha = 0.5$.
To efficiently train the computationally intensive models, we simultaneously employ VLLM~\cite{VLLM2023Woosuk}, DeepSpeed~\cite{ZERO2020RajbhandariRRH20} and Flash Attention~\cite{FlashAttention2023ABS230708691}. On 8 NVIDIA A800 80GB GPUs, the experiments on 1.5B, 3B, and 7B models take 8 hours, 10 hours, and 16 hours, respectively.

\subsection{Evaluation}

\paragraph{Baseline.} We employ DPO~\cite{DPO2023Rafailov}, KTO~\cite{KTO2024Kawin}, and Focused-DPO~\cite{FocusedDPO2025Kechi} as our baseline. For the DPO and KTO, we use the chat format to generate 5 responses for each problem. For Focused-DPO, we follow the paper to generate 10 responses and select the 5000 samples with the longest common prefix and suffix. The other pipeline and hyperparameters are the same as the implementation details of \name.

\begin{table*}[t]
\centering
\begin{tabular}{l|cc|cc|c|c|cc}
\toprule
\multirow{2.5}{*}{Method} & \multicolumn{2}{c|}{HumanEval} & \multicolumn{2}{c|}{MBPP} & \multirow{2.5}{*}{APPS} & \multirow{2.5}{*}{LiveCodeBench} & \multicolumn{2}{c}{BigCodeBench} \ \\ \cmidrule(lr){2-5} \cmidrule(lr){8-9}
~ & Base & Plus & Base & Plus & ~ & ~ & Full & Hard \\ \midrule
\name & 75.6 & 71.3 & \textbf{75.7} & \textbf{64.8} & \textbf{9.9} & \textbf{10.8} & \textbf{29.2} & \textbf{8.1} \\ \midrule 
~~w/ DPO & 74.4 & 69.5 & 73.5 & 63.8 & 9.3 & 7.8 & 28.6 & 6.8 \\
~~w/ DPO (Data Equal) & 66.5 & 59.8 & 71.4 & 59.0 & 2.0 & 4.1 & 25.9 & 6.1 \\ \midrule 
~~w/o AST & 75.0 & 69.5 & 74.1 & 62.4 & 9.9 & 9.7 & 28.8 & 3.4 \\ 
~~w/o Curriculum  & \textbf{78.7} & \textbf{73.2} & 73.8 & 64.3 & 9.8 & 9.0 & 28.3 & 7.4\\ 
~~w/ $\alpha = 0$ & 75.0 & 70.7 & 74.1 & 62.4 & 9.6 & 8.6 & 28.7 & 8.1 \\ 
~~w/ $\alpha = 1$ & 78.7 & 72.0 & 74.1 & 64.0 & 9.6 & 9.7 & 29.1 & 8.1 \\ \midrule
~~w/ suf loss & 77.4 & 70.7 & 75.4 & 64.6 & 9.8 & 8.2 & 29.0 & 7.4 \\ 
~~w/ pre \& suf loss & 76.2 & 70.7 & 74.9 & 64.6 & 9.7 & 8.1 & 29.1 & 7.4 \\
~~w/o suf & 73.2 & 69.5 & 75.1 & 62.4 & 9.8 & 9.7 & 29.1 & \textbf{8.1} \\
\bottomrule
\end{tabular}
\caption{Ablation results based on Qwen2.5-Coder-1.5B-Instruct. The score is Pass@1 accuracy.}
\label{tab:ablation}
\end{table*}

\paragraph{Results.} Table~\ref{tab:main} shows the pass@1 accuracy of different models on HumanEval, MBPP, APPS, LiveCodeBench, and BigCodeBench. Based on the results, we have the following findings:

(1) The proposed \name consistently outperforms standard DPO and KTO across all models on a variety of code-related tasks. Specifically, for Qwen2.5-Coder-1.5B-Instruct, \name surpasses DPO by an average of 1.5 points across all test sets. Similarly, for Qwen2.5-Coder-3B-Instruct and Qwen2.5-Coder-7B-Instruct, \name outperforms DPO by an average of 1.6 points across all test sets. The similar increase in different sizes demonstrates the effectiveness of the proposed method.

(2) The proposed \name shows a more significant improvement over DPO on most test sets compared to the APPS test set. For Qwen2.5-Coder-1.5B-Instruct, \name exceeds DPO by an average of 1.5 points on all test sets except for the APPS test set, where the improvement is only 0.6 points. This suggests that the proposed method enables the model to learn more diverse code patterns and achieve better generalization.

(3) The proposed \name demonstrates comparable performance to Focused-DPO on Qwen2.5-Coder-7B-Instruct, but achieves a significant improvement on Qwen2.5-Coder-1.5B-Instruct. This is because generations from the weaker model tend to have shorter comment prefixes and suffixes, which diminishes the effectiveness of Focused-DPO. In contrast, our method does not rely on aligning specific features of the generations, leading to consistent improvements across all three models.

\subsection{Detailed Analysis}

\subsubsection{Ablation Study}

Here, we check how each component contributes to the final performance. We prepare two group variants of our method based on Qwen2.5-Coder-1.5B-Instruct:
(1) The first group is related to DPO. \underline{w/ DPO} denotes constructing training data w/o FIM. \underline{w/ DPO (Data Equal)} denotes training with more epochs to ensure the total training samples are the same as \name.
(2) The second group is related to the proposed components. \underline{w/ AST} denotes selecting target code snippets randomly. \underline{w/ Curriculum} denotes training with random order. \underline{w/ $\alpha = 0$} denotes training only with the standard format. \underline{w/ $\alpha = 1$} denotes training only with the FIM format. 

Table~\ref{tab:ablation} presents the pass@1 accuracy for various model variants. Based on the results, we have the following findings:

(1) As shown, all variants perform worse than \name across most test sets, except for HumanEval. These results suggest that all components~(\ie AST enhanced FIM, curriculum learning, and mixing FIM training format and Chat format) are critical for enhancing performance. 

(2) Furthermore, \underline{w/ DPO (Data Equal)} performs worse than \underline{w/ DPO}, indicating that DPO benefits more from diverse, large-scale training data than from simply increasing the number of epochs. Increasing epochs can lead to overfitting. In contrast, \name effectively generates more diverse training data through FIM, resulting in improved performance. 

(3) Lastly, \underline{w/ Curriculum} and \underline{w/ $\alpha = 1$} outperform \name on HumanEval. A potential explanation for this is that the HumanEval test cases differ significantly from those in the APPS, making fine-tuning with longer code during the final stage detrimental to HumanEval performance.

\subsubsection{Effect of Loss on Prefix and Suffix}

As we analyzed in Section~\ref{sec:pre}, the DPO loss on the prefix segment does not impact the optimization, but the DPO loss on the suffix segment harms the optimization. Here, we conduct experiments to check the effect of loss on the prefix segment and the suffix segment. Specifically, we prepare three variants based on Qwen2.5-Coder-1.5B-Instruct. \underline{w/ suf loss} computes the loss on both the middle and the suffix segments. \underline{w/ pre \& suf loss} computes the loss on all segments. \underline{w/o suf} remove suffix in the whole pipeline, \aka, we only use the query and the prefix to prompt the model to generate various generations, effectively setting the suffix to be empty.

Table~\ref{tab:ablation} presents the pass@1 accuracy for the variants. 
As shown in the table, \underline{w/ suf loss} significantly performs worse than \name on LiveCodeBench, which is the most difficult test set. This indicates that the DPO loss on the suffix segment hurts the optimization.
However, \underline{w/ suf loss} and \underline{w/ pre \& suf loss} perform similarly. This indicates that the DPO loss on the prefix segment does not impact the optimization.
This phenomenon is less noticeable on other datasets because they are simpler. The difficulty of LiveCodeBench highlights the impact of the DPO loss on the suffix segment, while simpler datasets don’t show the same effect.
Finally, \underline{w/o suf} performs worse than StructureCoder in most settings. This is because concatenating the middle and suffix spans results in overly long input sequences, which may hinder the model’s ability to focus on critical code blocks. This observation aligns with our theoretical insights discussed in Section~\ref{sec:weak_of_dpo}.
\section{Related Work} \label{sec:rel}

\subsection{Large Language Models for Code}

Previous works have introduced LLMs for the code domain. 
OpenAI introduced Codex~\cite{CodexABS210703374}, and Google introduced PaLM-Coder~\cite{PaLM2023ChowdheryNDBMRBCSGSSTMRBTSPRDHPBAI23}. 
There are also several open-source LLMs for the code domain, such as CodeGen~\cite{CodeGen2023NijkampPHTWZSX23}, Incoder~\cite{InCoderFriedAL0WSZYZL23}, SantaCoder~\cite{SantaCoderABS230103988}, StarCoder~\cite{StarCoderABS230506161}, StarCoder-2~\cite{StarCoder22024ABS240219173}, CodeGeeX~\cite{CodeGeeX2023ZhengXZDWXSW0LS23}, Code Llama~\cite{CodeLlamaABS230812950}, and DeepSeek-Coder~\cite{DeepSeekCoderABS240114196}. 
Recently, Deepseek-Coder-V2~\cite{DeepseekCoderV22024DeepSeek} and Qwen2.5-Coder~\cite{QwenCoder2024Binyuan} have been proposed, which achieve performance close to that of closed-source models.

Instruction tuning is demonstrated to be an effective method to improve the ability of LLMs in specific domains. Code Alpaca~\cite{CodeAlpaca2023GitHub} applied Self-Instruct~\cite{SelfInstruct2022WeiBZGYLDDL22} to fine-tune LLMs with ChatGPT-generated instructions. 
WizardCoder~\cite{WizardCoder2023ABS230608568} proposed Code Evol-Instruct, which evolves Code Alpaca data using the ChatGPT to generate more complex and diverse datasets. MagiCoder~\cite{Magicoder2023ABS231202120}, WaveCoder~\cite{WaveCoderABS231214187}, and InverseCoder~\cite{InverseCoder2024Yutong} proposed some methods to make full use of source code. OpenCodeInterpreter~\cite{OpenCodeInterpreter2024ABS240214658}, AutoCoder~\cite{AutoCoder2024Bin} and ReflectionCoder~\cite{ReflectionCoder2024Ren} proposed to utilize a compiler to enhance code LLMs.

\subsection{Alignment for Code}

Reinforcement learning-based alignment methods have been shown to improve LLM performance. DPO~\cite{DPO2023Rafailov}, though widely adopted, provides limited gains in code generation tasks~\cite{DPOAnalysis2024Xu}. PLUM~\cite{PLUM2024Zhang} uses GPT-4 to generate test cases for code validation and ranking. CodeDPO~\cite{CodeDPO2024Zhang} applies a PageRank-inspired algorithm for iterative preference scoring. StepCoder~\cite{StepCoder2024ABS240201391} employs PPO~\cite{PPO2017Schulman} to optimize general code generation, and CodeOptimise~\cite{CodeOptimize2024Leonidas} focuses on improving runtime efficiency. Concurrent with our work, Focused-DPO~\cite{FocusedDPO2025Kechi} also identified the issue of common prefixes and suffixes. However, their approach heavily depends on model generations that contain these patterns, whereas our method eliminates such dependency.

Unlike previous DPO-based approaches to code generation, our approach focuses on maximizing the utility of limited data to construct more effective preference pairs. While existing methods often rely on larger datasets or complex augmentation strategies, we take a data-efficient approach by making the most of even small-scale datasets. By introducing FIM~\cite{FIM2022ABS220714255}, we can efficiently extract high-quality preference pairs, significantly improving the alignment of code generation models while utilizing a limited amount of labeled data.
\section{Conclusion} \label{sec:con}

In this paper, we introduced a novel approach to enhance code generation performance for LLMs using the fill-in-the-middle~(FIM). By generating fine-grained DPO pairs from limited test cases and employing a curriculum training method, we maximized the utility of available data, leading to improved model performance. Our experiments on benchmarks like HumanEval~(+), MBPP~(+), APPS, LiveCodeBench, and BigCodeBench demonstrated the effectiveness of our approach in code-related tasks. Future work could explore further refinements, such as alternative segmentation strategies or additional data sources.

\section*{Limitations} 

One key limitation of our approach is that it requires the model to possess strong Fill-In-the-Middle~(FIM) capabilities, which are essential for effectively generating fine-grained DPO pairs. However, most current models, after supervised fine-tuning, may struggle to maintain high FIM performance. As a result, the effectiveness of our method is closely tied to the underlying model’s proficiency in FIM, a capability that is not yet universally available in existing LLMs. This reliance on FIM limits the applicability of our approach to models that have been specifically optimized for this task.
Another limitation is that our method is currently focused solely on the code generation domain. While we believe that similar phenomena may exist in other domains, our approach has not yet been tested outside the code generation field. Further research would be required to explore the transferability to other closed-question tasks.

\section*{Ethics Statement}

The model utilized in this paper, Qwen2.5-Coder~\cite{StarCoderABS230506161}, is licensed for academic research purposes\footnote{\url{https://github.com/QwenLM/Qwen2.5-Coder}}. Furthermore, the data employed in this study, APPS~\cite{APPS2021Dan}, is also licensed for academic research purposes\footnote{\url{https://github.com/hendrycks/apps}}. 

\section*{Acknowledgment}

This project is funded in part by National Key R\&D Program of China Project 2022ZD0161100, by the Centre for Perceptual and Interactive Intelligence (CPII) Ltd under the Innovation and Technology Commission (ITC)’s InnoHK, and in part by NSFC-RGC Project N\_CUHK498/24. 

\bibliography{
    bibs/code_llm, 
    bibs/code_sft, 
    bibs/general_llm, 
    bibs/other
}

\clearpage\appendix\section*{Appendix}

\section{Dataset Construction}

To evaluate the structural diversity of training data constructed for our method, we performed a statistical analysis of code blocks extracted via Abstract Syntax Tree (AST) parsing. Specifically, we retained four categories of syntactic nodes as middle targets for the FIM task: if, for, while, and def (function definitions). The figure below presents the distribution of these node types in the final training dataset.

\begin{figure}[t]
    \centering
    \includegraphics[width=\columnwidth]{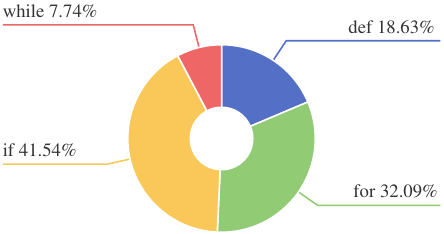}
    \caption{Distribution of extracted code blocks based on AST node types. This reflects the syntactic diversity in the training corpus and validates the representativeness of selected blocks.}
    \label{fig:stat}
\end{figure}

As shown in Figure~\ref{fig:stat}, among all segments used in FIM training, if blocks constitute the largest proportion (41.54\%), followed by for loops (32.09\%), def (function) blocks (18.63\%), and while loops (7.74\%). This distribution indicates that the selected code snippets exhibit substantial structural variety, with conditional~(if) and iterative~(for and while) constructs dominating the sample pool. Such a mix helps the model learn frequent control flow patterns more effectively in FIM tasks. Although while and def blocks appear less frequently, they still contribute essential control and functional semantics, ensuring the structural completeness of the training examples.

To construct the training dataset, we follow a systematic pipeline that parses source code into abstract syntax trees, generates fill-in-the-middle (FIM) prompts, evaluates model completions against test cases, and derives preference pairs from completions. The detailed procedure is outlined in Algorithm~\ref{alg:data}.

\section{Insights of \name}

\subsection{Benefits of FIM}

Fill-in-the-Middle (FIM)~\cite{FIM2022ABS220714255} has been shown to preserve the model's ability to generate code effectively~\cite{StarCoderABS230506161, CodeLlamaABS230812950, FIMSE2025Houxing}, while offering several distinct advantages that enhance overall model performance and robustness. By requiring generation conditioned on both preceding and succeeding context, FIM strengthens contextual understanding and promotes deeper semantic reasoning. Rather than replacing standard code generation, FIM acts as a powerful complement—particularly well-suited for practical scenarios such as code refactoring, patching, and bug fixing. Its bidirectional formulation improves generalization and equips models to handle complex code dependencies more effectively. Our empirical results further demonstrate that FIM-based training leads to notable gains in code generation quality, underscoring its value as a targeted and practical enhancement to existing methodologies.

\subsection{Comparison with Step-DPO}

While our proposed method shares some similarities with Step-DPO regarding loss construction and data generation, it offers distinct improvements in the following aspects:

(1) Step-DPO relies on labeling intermediate reasoning steps either through human annotation, which is prohibitively expensive, or through automated labeling methods such as GPT-4, which introduces potential biases as highlighted by recent studies~\cite{deepseekr12025deepseek}. In contrast, our proposed method utilizes FIM to automatically generate and assess intermediate reasoning fragments based on objective evaluation criteria from test samples. By eliminating the reliance on costly human annotation and avoiding the biases associated with automated labeling tools, our method offers an unbiased alternative for training and evaluating reasoning capabilities in large language models.

(2) Annotating individual steps in Step-DPO can be inherently ambiguous, especially for long Chain-of-Thought (CoT). An incorrect intermediate step can sometimes contribute meaningfully to reaching a correct final answer — for example, models like o1 often make a mistake in early reasoning steps, such as miscalculating a quantity or misunderstanding part of the problem, but then later detect the inconsistency in their own logic and correct for it in the final answer. In such cases, although the intermediate step is technically incorrect, it still plays a functional role in producing the right outcome. This makes it difficult to definitively label individual steps as preferred or dispreferred. Our approach overcomes this issue by fixing the suffix of the code, requiring the generated intermediate code (the middle segment) to coherently bridge the given prefix and suffix. This design fundamentally eliminates ambiguity in annotation, clearly delineating preferred and dispreferred middle segments.

\section{Additional Experiment}

\subsection{FIM Evaluaion}

To thoroughly assess the Fill-In-the-Middle (FIM) capabilities of our proposed method, we conducted a series of targeted evaluations using the APPS benchmark. \nocite{zong2024mova, ma2024exploring, zong2024easyref, jiang2025t2i, quick2022ren, lestm2022ren} Specifically, we curated a set of 22,044 samples by randomly masking contiguous code spans from the reference solutions. The models were then tasked with accurately reconstructing these masked segments solely based on the surrounding context, thereby simulating realistic code completion scenarios.

As summarized in Table~\ref{tab:fim}, the application of our method consistently improves FIM performance across all evaluated model scales. Notably, our approach achieves the highest Pass@1 accuracy in each model category, surpassing several strong baselines. \nocite{ lu2024mathgenie, lu2024mathcoder2, lu2025webgenbench, wang2023mathcoder, wang2025mathcodervl, wang2024mathvision} These findings underscore the robustness and generalizability of our method in enhancing middle-span code generation, demonstrating its potential as a principled strategy for strengthening contextual reasoning in large language models for programming tasks.

\begin{table}[t]
\centering
\begin{tabular}{l|c}
\toprule
\textbf{Model} & \textbf{FIM Pass@1} \\
\midrule
Qwen-2.5-Coder-1.5B-Inst & 27.1 \\
\quad w/ DPO & 27.5 \\
\quad w/ KTO & 27.4 \\
\quad w/ Focused-DPO & 27.4 \\
\quad w/ \name & \textbf{28.6} \\
\midrule
Qwen-2.5-Coder-3B-Inst & 31.1 \\
\quad w/ DPO & 31.5 \\
\quad w/ KTO & 31.6 \\
\quad w/ Focused-DPO & 31.7 \\
\quad w/ \name & \textbf{32.9} \\
\midrule
Qwen-2.5-Coder-7B-Inst & 35.7 \\
\quad w/ DPO & 35.6 \\
\quad w/ KTO & 35.9 \\
\quad w/ Focused-DPO & 36.1 \\
\quad w/ \name & \textbf{36.7} \\
\bottomrule
\end{tabular}
\caption{Pass@1 accuracy on the FIM task.}
\label{tab:fim}
\end{table}

\begin{algorithm*}[h]

\caption{Training Data Construction Pipeline}

\KwIn{$D = \{(q, c, t)\}$, where $q$ is a question, $c$ is the corresponding code, and $t$ are test cases}
\KwOut{Final training set ready for fine-tuning}

\ForEach{$(q, c, t) \in D$}{
    Parse $c$ into an Abstract Syntax Tree (AST)\;
    Extract target code blocks $M = \{m_1, m_2, \ldots, m_n\}$\;

    \ForEach{$m_i \in M$}{
        Extract prefix $p_i$ and suffix $s_i$\;
        Construct FIM prompt: \texttt{<PRE>} Convert($q$) $p_i$ \texttt{<SUF>} $s_i$ \texttt{<MID>}\;
        Generate $m$ completions $G_i = \{g_i^{(1)}, \ldots, g_i^{(m)}\}$\;

        \ForEach{$g_i^{(j)} \in G_i$}{
            $full\_code \gets p_i + g_i^{(j)} + s_i$\;
            Execute $full\_code$ on test cases $t$\;
            \eIf{passes}{
                Add $g_i^{(j)}$ to correct candidates\;
            }{
                Add $g_i^{(j)}$ to incorrect candidates\;
            }
        }
    }

    \ForEach{incorrect completion $g^{-}$}{
        Find closest correct $g^{+}$ by edit distance\;
        Form preference pair $(g^{+}, g^{-})$\;
    }
}

\ForEach{preference pair $(g^{+}, g^{-})$}{
    Sample prompt format using Bernoulli distribution ($\alpha$)\;
    \eIf{format $=$ FIM}{
        Positive $\gets$ \texttt{<PRE>} Convert($q$) $p_i$ \texttt{<SUF>} $s_i$ \texttt{<MID>} $g^{+}$\;
        Negative $\gets$ \texttt{<PRE>} Convert($q$) $p_i$ \texttt{<SUF>} $s_i$ \texttt{<MID>} $g^{-}$\;
    }{
        Positive $\gets$ User: $q$; Assistant: $p_i + g^{+} + s_i$\;
        Negative $\gets$ User: $q$; Assistant: $p_i + g^{-} + s_i$\;
    }
    Add sample to training set\;
}

Sort training samples by length of $g^{+}$ (short $\to$ long)\;
\Return Final training set\;

\label{alg:data}
\end{algorithm*}

\subsection{Case Study}

Here, we use a case from Qwen2.5-Coder-1.5B-Instruct, Qwen2.5-Coder-3B-Instruct and Qwen2.5-Coder-7B-Instruct to show the effectiveness of the proposed method. Specifically, we compute the DPO reward~($r(s, a) = \beta \log \pi_{\theta} (s|a) - \beta \log \pi_{ref} (s|a)$) for each token in both correct and incorrect responses. The DPO is used as a baseline for comparison.

\begin{figure*}[t]
    \centering
    \subfigure[Credit assignment Qwen2.5-Coder-1.5B-Instruct w/ DPO.]{ \label{fig:1b_dpo}
        \includegraphics[width=0.9\textwidth]{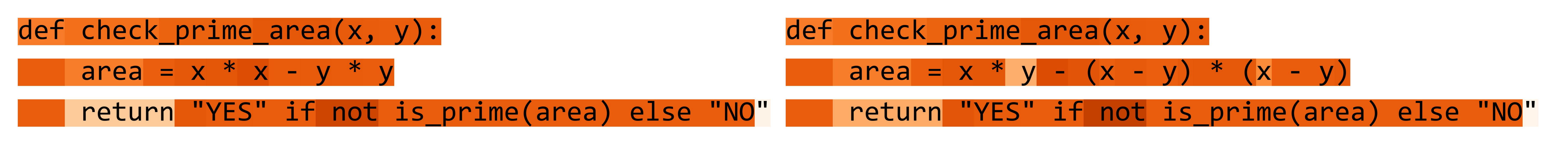}
    }
    \subfigure[Credit assignment Qwen2.5-Coder-1.5B-Instruct w/ \name.]{ \label{fig:1b_our}
        \includegraphics[width=0.9\textwidth]{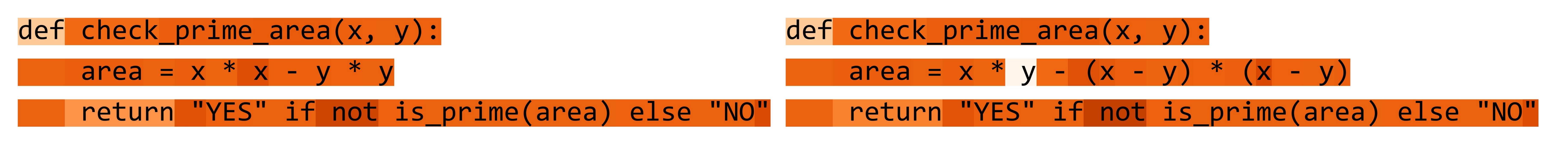}
    }
    \caption{Credit assignment with different methods. Due to the limited space, we omit the previous function, only keeping the key function. The left is the correct response and the right is the incorrect response (error is introduced in Line 2). Each token is colored corresponding to the DPO implicit reward (darker is higher).}
    \label{fig:1b_case}
\end{figure*}

\begin{figure*}[t]
    \centering
    \subfigure[Credit assignment Qwen2.5-Coder-3B-Instruct w/ DPO.]{ \label{fig:3b_dpo}
        \includegraphics[width=0.9\textwidth]{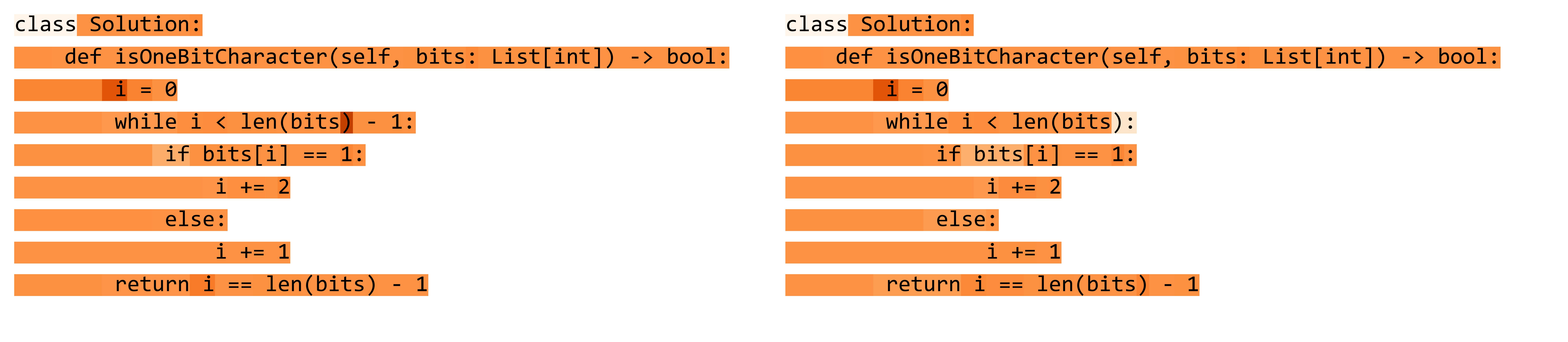}
    }
    \subfigure[Credit assignment Qwen2.5-Coder-3B-Instruct w/ \name.]{ \label{fig:3b_our}
        \includegraphics[width=0.9\textwidth]{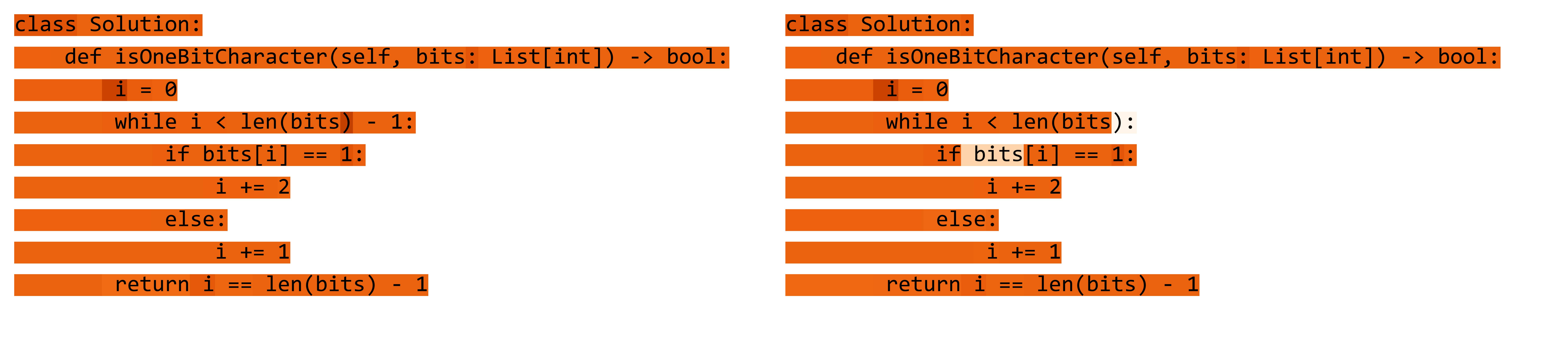}
    }
    \caption{Credit assignment with different methods on Qwen2.5-Coder-3B-Instruct. The left is the correct response and the right is the incorrect response (error is introduced from the last token of Line 4). Each token is colored corresponding to the DPO implicit reward (darker is higher).}
    \label{fig:3b_case}
\end{figure*}

\begin{figure*}[t]
    \centering
    \subfigure[Credit assignment Qwen2.5-Coder-7B-Instruct w/ DPO.]{ \label{fig:7b_dpo}
        \includegraphics[width=0.9\textwidth]{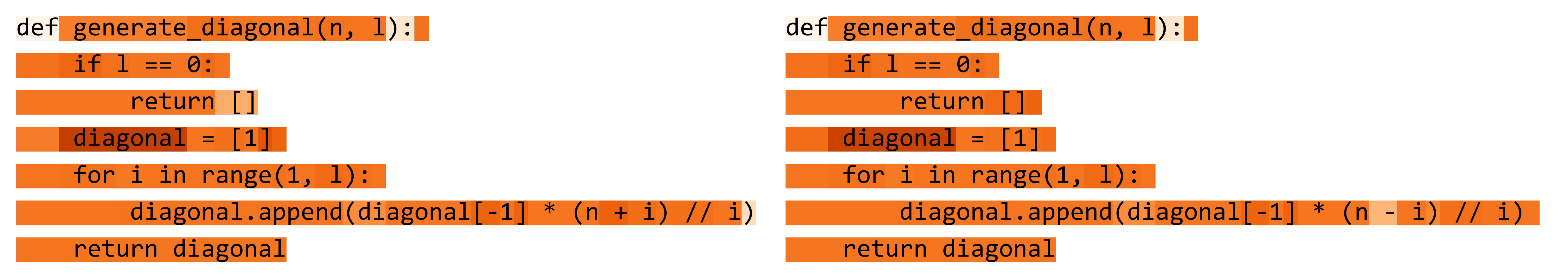}
    }
    \subfigure[Credit assignment Qwen2.5-Coder-7B-Instruct w/ \name.]{ \label{fig:7b_our}
        \includegraphics[width=0.9\textwidth]{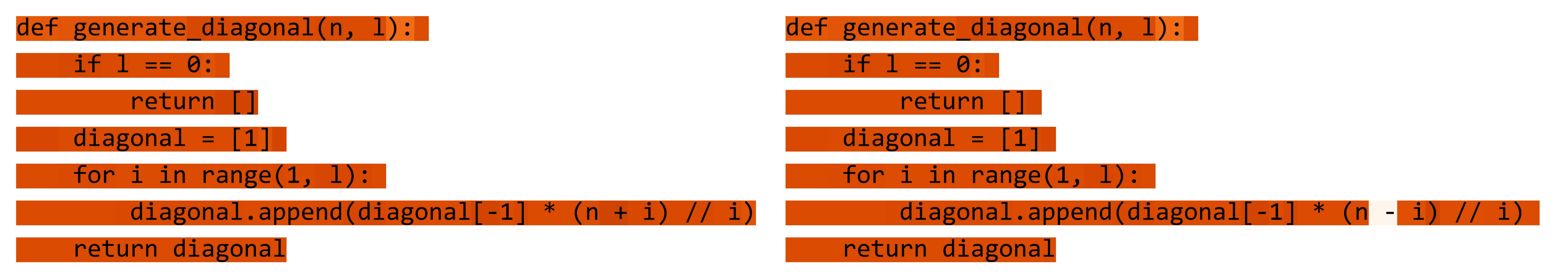}
    }
    \caption{Credit assignment with different methods on Qwen2.5-Coder-7B-Instruct. The left is the correct response and the right is the incorrect response (error is introduced in Line 6). Each token is colored corresponding to the DPO implicit reward (darker is higher).}
    \label{fig:7b_case}
\end{figure*}

As shown in Figure~\ref{fig:1b_case}, the error is introduced in Line 2, and the proposed method, \name, successfully identifies the token corresponding to the erroneous statement ($x * x$ and $x * y$). For Qwen2.5-Coder-3B-Instruct, as shown in Figure~\ref{fig:3b_case}, the proposed \name successfully assigns high reward to the token corresponding to the erroneous statement ($len(bits) - 1$ and $len(bits)$). For Qwen2.5-Coder-7B-Instruct, as shown in Figure~\ref{fig:7b_case}, the proposed \name effectively highlights he erroneous statement ($n + i$ and $n - i$). These demonstrate the ability of our method to localize specific errors and focus attention on the relevant details. On the other hand, the DPO fails to detect these errors, highlighting its limitation in handling errors with high precision, especially when the training data is sparse. While the DPO might perform well with abundant training examples, its performance drops in the presence of limited data, making it less effective for tasks requiring fine-grained error detection in these conditions.

\end{document}